\documentclass[conference]{ieeeconf}
\IEEEoverridecommandlockouts
\usepackage[utf8]{inputenc}
\usepackage{comment}
\usepackage{graphicx}
\usepackage{xcolor}
\usepackage{wrapfig}
\usepackage{lipsum}
\definecolor{rred}{RGB}{169, 50, 38}
\definecolor{ggreen}{RGB}{34, 153, 84}
\definecolor{bblue}{RGB}{36, 113, 163}
\definecolor{ppurple}{RGB}{125, 60, 152}
\definecolor{yyellow}{RGB}{214, 137, 16}
\definecolor{ggrey}{RGB}{112, 123, 124}

\usepackage{amsmath}
\usepackage{amssymb}
\usepackage{amsthm}
\theoremstyle{definition}
\usepackage{tikz}
\usetikzlibrary{automata,positioning,angles,quotes}
\usepackage{tikz-cd}

\newtheoremstyle{main}
{1em} 
{1em} 
{\normalfont} 
{0pt} 
{\scshape} 
{\\*} 
{2pt} 
{\thmname{#1}\thmnumber{ #2}: \thmnote{\itshape #3}} 

\makeatletter
\let\NAT@parse\undefined
\makeatother
\usepackage{cite}
\usepackage[pdfa,colorlinks,bookmarks=true,bookmarksopen,bookmarksnumbered,allcolors=ggreen]{hyperref}

\usepackage[algoruled,vlined,linesnumbered]{algorithm2e}

\SetAlgoSkip{SkipBeforeAndAfter}

\SetCommentSty{mycommfont}
\SetKw{Continue}{continue}
\SetAlgorithmName{Algorithm}{Alg.}

\usepackage{booktabs}
\usepackage{multirow}
\usepackage{multicol}
\usepackage{makecell}

\usepackage[font=footnotesize]{caption}
\usepackage[font=footnotesize]{subcaption}
\usepackage[export]{adjustbox}

\usepackage[
activate   = {true},
protrusion = true,
expansion  = true,
kerning    = true,
spacing    = true,
tracking   = false,
auto       = true,
selected   = true,
factor     = 1000,
stretch    = 20,
shrink     = 20,
]{microtype}

\usepackage{xspace}

\newcommand{\method}{\textsc{ActivePusher}\xspace}
\newcommand{\NTK}{\textsc{NTK}\xspace}
\newcommand{\BAIT}{\textsc{BAIT}\xspace}
\newcommand{\sst}{\textsc{SST}\xspace}

\newcommand{\GP}{\textsc{GP}\xspace}
\newcommand{\set}{\mathcal{S}}

\newcommand{\R}[1]{{\ensuremath{\mathbb{R}^{#1}}}\xspace}
\newcommand{\SEThree}{\ensuremath{SE(3)}\xspace}
\newcommand{\SETwo}{\ensuremath{SE(2)}\xspace}

\newcommand{\SOTwo}{\ensuremath{SO(2)}\xspace}

\newcommand{\X}{\ensuremath{\mathcal{X}}\xspace}

\newcommand{\U}{\ensuremath{\mathcal{U}}\xspace}
\newcommand{\uu}{u}
\newcommand{\Xfree}{\ensuremath{\X_{\text{free}}}\xspace}
\newcommand{\Xobs}{\ensuremath{\X_{\text{obs}}}\xspace}
\newcommand{\x}{x}
\newcommand{\dof}{\textsc{dof}\xspace}

\newcommand{\T}{{\ensuremath{\mathcal{T}}}\xspace}

\usepackage{amssymb}

\title{\LARGE \bf
\method: Active Learning and Planning \\ with Residual Physics for Nonprehensile Manipulation
}
\author{
  Zhuoyun Zhong, Seyedali Golestaneh, and Constantinos Chamzas
  \thanks{%
  This work was supported in part by Amazon WPI RBE Research Award 2025, NSF CRII Grant No. 2451108 and WPI funds. All authors are affiliated with the Department of Robotics Engineering, Worcester Polytechnic Institute (WPI), Worcester, MA 01609, USA {\tt\small \{zzhong3, sgolestaneh, cchamzas\} @ wpi.edu}
  }
}

\usepackage{draftwatermark}
\SetWatermarkText{\large Accepted for publication in the 2026 IEEE International Conference on Robotics \& Automation (ICRA 2026)}
\SetWatermarkColor[gray]{0.5}
\SetWatermarkFontSize{0.42cm}
\SetWatermarkAngle{0}
\SetWatermarkHorCenter{0.5\paperwidth}
\SetWatermarkVerCenter{0.5in}

\begin{document}

\maketitle
\thispagestyle{empty}
\pagestyle{empty}


\renewcommand\twocolumn[1][]{#1}%
\maketitle

\begin{abstract}
Planning with learned dynamics models offers a promising approach toward versatile real-world manipulation, particularly in nonprehensile settings such as pushing or rolling, where accurate analytical models are difficult to obtain.
However, collecting training data for learning-based methods can be costly and inefficient, as it often relies on randomly sampled interactions that are not necessarily the most informative.
Furthermore, learned models tend to exhibit high uncertainty in underexplored regions of the skill space, undermining the reliability of long-horizon planning.
To address these challenges, we propose \method, a novel framework that combines residual-physics modeling with uncertainty-based active learning, to focus data acquisition on the most informative skill parameters.
Additionally, \method seamlessly integrates with model-based kinodynamic planners, leveraging uncertainty estimates to bias control sampling toward more reliable actions.
We evaluate our approach in both simulation and real-world environments, and demonstrate that it consistently improves data efficiency and achieves higher planning success rates in comparison to baseline methods.
The source code is available at \url{https://github.com/elpis-lab/ActivePusher}.
\end{abstract}


\section{Introduction}
\label{sec:introduction}

Model-based planning methods offer a powerful framework for generalizing robotic behavior and enabling long-horizon decision making~\cite{orthey2024-review-sampling}. Such methods often require a predictive model of the system’s dynamics, especially in kinodynamic planning where dynamic constraints must be considered. The effectiveness of these approaches thus critically depends on the accuracy of the underlying forward dynamics model. Inaccuracies in this model can cause cascading errors during execution, particularly in contact-rich settings such as nonprehensile manipulation (e.g., pushing and rolling), where even minor deviations in predicted trajectories may accumulate and lead to irrecoverable task failure.

Accurately modeling the dynamics for these tasks is challenging. Analytical physics-based models often rely on simplified assumptions about friction, contact geometry, and mass distribution, making them brittle in practice~\cite{mason2018toward}. As an alternative, data-driven approaches can learn dynamics directly from interaction data, either from scratch or by refining simplified analytical models through residual learning~\mbox{\cite{tossingbot, residual-pushing}}. However, these methods face two key limitations in real-world robotic settings: 

\begin{itemize} 
\item \textbf{Sample inefficiency:} Learning accurate models often requires large amounts of interaction data, which is costly and time-consuming to collect on physical systems. 

\begin{figure}[ht!]
    \centering
    \includegraphics[width=0.9\linewidth]{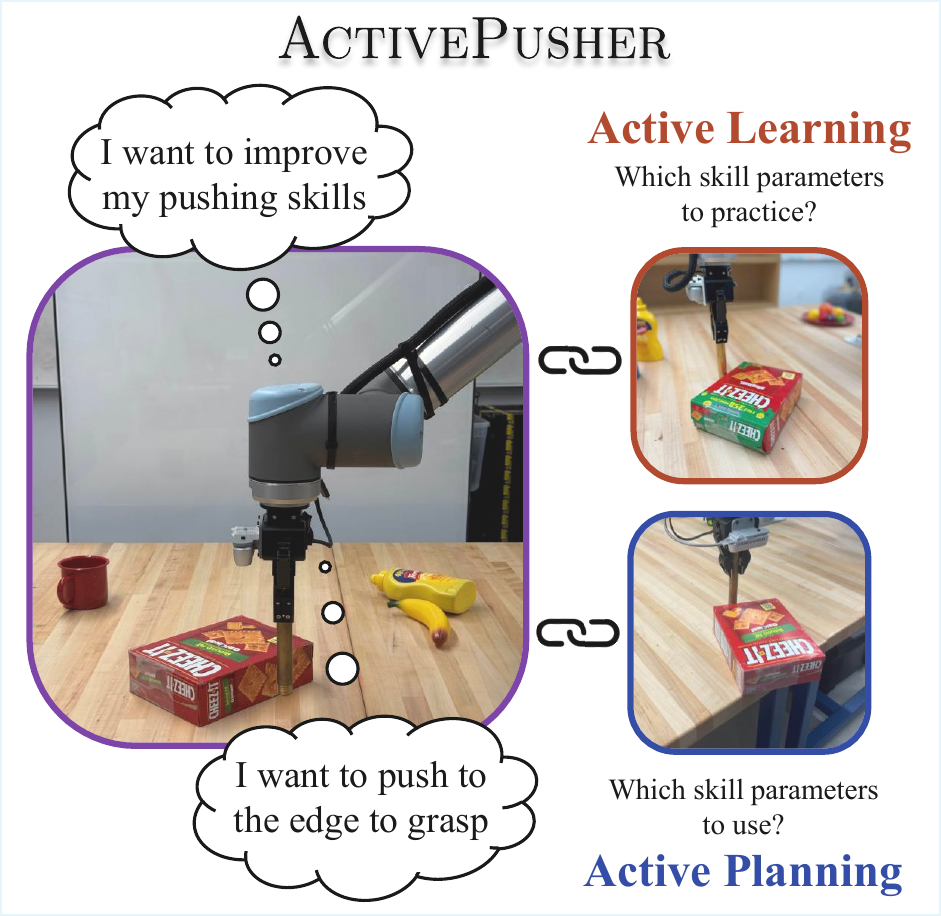}
    \caption{Two key challenges addressed by \method. During learning, the robot must choose the most informative skill parameters to efficiently improve its skills (Active Learning). When planning, the robot should select skill parameters with low model uncertainty to ensure reliable task completion (Active Planning).} 
    \label{fig:intro}
    \vspace{-2ex}
\end{figure}

\item \textbf{Inaccuracy in underexplored regions:} Models may perform poorly in sparsely explored regions of the skill space, leading to unreliable predictions and large deviation from plans during execution. 
\end{itemize}

In this paper, we propose \method, a framework that tightly integrates residual physics, active learning, and active kinodynamic planning, to address both challenges in the context of nonprehensile pushing, illustrated in \autoref{fig:intro}. The core idea is to quantify epistemic uncertainty in a learned neural network model with residual physics. This uncertainty estimate allows the system to actively target informative skill parameters for learning and reliable ones for planning.

To estimate the epistemic uncertainty of a neural network model, we leverage the Neural Tangent Kernel (\NTK)~\cite{ntk}. 
During learning, rather than sampling pushing actions at random to practice, \method actively queries the \NTK for uncertainty estimation and identifies actions with the highest expected information gain using Batch Active learning via Information maTrices (\BAIT) strategy~\cite{bait}. This targeted exploration enables the model to improve rapidly with far fewer interactions.
During planning, the uncertainty estimates are incorporated into a sampling-based kinodynamic planner~\cite{orthey2024-review-sampling}, biasing action sampling toward high-confidence skill parameters to maximize task success.

By focusing on where the model is uncertain to learn, and where the model is certain to act, our approach tightly integrates learning and planning, enabling robust nonprehensile manipulation with few real-world interactions per task. Crucially, \method operates without high-fidelity simulation, large offline datasets, or human demonstrations.
Our main contributions are:
\begin{itemize}
\item  \textbf{Active learning of skill models}. We introduce a principled framework for data-efficient nonprehensile skill learning by selecting skill parameters that maximize expected information gain, enabling data collection in the most informative manner.
\item \textbf{Active uncertainty-aware planning}. We propose a novel planning strategy that integrates model uncertainty into an asymptotically optimal kinodynamic planner, guiding action sampling toward reliable actions and improving overall task success rate.
\item \textbf{Empirical validation in simulation and the real world}. We demonstrate the effectiveness of our approach with multiple objects and nonprehensile manipulation tasks, showing significantly improved data efficiency and planning success over baselines.
\end{itemize}

\section{Related Work}
\textsc{ActivePusher} draws ideas from several areas, such as residual learning, active learning, and kinodynamic planning. Here, we briefly review each of these areas in the context of nonprehensile manipulation, with a focus on pushing.
\textbf{Residual Model Learning} combines the strengths of analytical and data-driven approaches by training a neural network to predict corrections to an approximate physics model. 
In robotic manipulation, analytical models and simulations can offer useful priors but are often coarse approximations of real dynamics, sensitive to assumptions on physical parameters~\cite{hogan2020reactive}. Conversely, fully data-driven methods~\cite{poking} can model complex behaviors but require large amounts of real-world data.
Residual learning reduces this burden by modeling only the error between physics and reality, improving data efficiency and real-world performance~\cite{tossingbot, residual-pushing}. Building on this paradigm, \textsc{ActivePusher} further improves data efficiency by actively selecting informative data points for refinement.
\textbf{Active Learning} is a well-established topic in machine learning that improves data efficiency by actively selecting which data points to label with model uncertainty~\mbox{\cite{bald, bait, activeregression}}. This paradigm naturally aligns with self-supervised robotic learning settings, where the data collection is expensive, and the robot should autonomously choose which interactions to collect. 
Several robotic learning methods have leveraged active learning to improve learning efficiency~\cite{activeleraningcontrol}.
In the context of skill learning, recent approaches~\cite{ltamp, lagrassa2024task} have applied active learning to accelerate skill learning with binary outcome using Gaussian Process (\GP). 
Uncertainty estimates from such models have also been used to provide robustness in control synthesis.~\cite{dkl-control}.
However, prior active skill learning methods focus primarily on binary success classification, with limited attention to regression-based skill models using neural networks. In contrast, we apply active learning in training predictive skill models with neural networks, and enable their integration into kinodynamic planners.
\textbf{Nonprehensile Manipulation}, particularly planar pushing problems, has been addressed by many approaches such as Model Predictive Control (MPC) and reinforcement learning (RL). RL methods~\cite{mujocoplayground, zhou2023hacman} can acquire complex behaviors, but require large amount of interaction data and often fail under distribution shift and sim-to-real transfer. MPC approaches that embed learned or physics models~\mbox{\cite{bauza2018data, hogan2020reactive, wang2024uno}} provide robust execution but require high-frequency state observation and control for closed-loop correction. Meanwhile, both methods remain inherently myopic and prone to local minima. 
On the other hand, kinodynamic motion planning methods provide global reasoning and can be generalized to complex environments~\cite{ren-object-centric, ren2023kinodynamic}. However, previous works often rely on analytical models, assume regular-shaped objects, and handle execution errors through online replanning~\cite{ren2023kinodynamic}, which cannot always recover from failure.
Our approach instead leverages a learned dynamics model and its uncertainty estimates to actively select reliable actions to build plans, enabling robust plans from the beginning.
\section{Problem Statement}
\label{sec:problem}

\textbf{Kinodynamic Planning:}
Let $\x \in \X$ denote the state and state-space, and $\uu \in \U$ denote the control and control-space of a robotic system~\cite{orthey2024-review-sampling}. We consider a dynamic system that follows time-invariant differential equations:
\begin{equation} 
\label{eq:motion}
\dot{x}(t) = f^*(\x(t), \uu(t))
\end{equation}
where $f^*$ is the true (unknown) forward dynamics model of the system.
Let $\Xobs \subset \X$ denote the obstacle (invalid) state space, and $\Xfree = \X \setminus \Xobs$ denote the free (valid) space. The start state is $\x_{\text{start}} \in \Xfree$, and the goal region is $X_{\text{goal}} \subseteq \Xfree$. 
The \textit{kinodynamic planning} problem is to determine a control duration $\T$ and a control function $\uu: [0, \T] \rightarrow \U$ such that the resulting trajectory satisfies $\x(0) = \x_{\text{start}}$, $\x(\T) \in X_{\text{goal}}$, and $\x(t) \in \Xfree$ for all $t \in [0, \T]$.

In this work, we focus on a planar pushing task, modeled as kinodynamic planning. We treat the state of the manipulated object as the system state and adopt an object-centric isotropy assumption, where the effect of a push is invariant to the object’s current pose. In addition, we assume that a static equilibrium condition is achieved after each action. Formally, this reduces the system dynamics to:
\begin{equation} 
    \dot{\x}(t) = f^*(\uu(t))
\end{equation} 

\textbf{Active Learning of Forward Dynamics Models:}
We define an interaction as the application of a control and observation of the associated states. Under the isotropic assumption, the interaction is simplified to ($\uu$, $\dot{\x}$). A set of applied controls and observed state changes forms a dataset for model learning.

The \textit{active learning of forward dynamics models} problem is to approximate the unknown dynamics model $f^*$ with a learned model $f$, such that it predicts the outcomes of applied controls with high predictive accuracy, while minimizing the number of interactions required for training.

\textbf{Active Kinodynamic Planning:}
Inspired by the concept of active learning with uncertainty quantification~\cite{bald, bait}, we define \textit{active kinodynamic planning} as kinodynamic planning that explicitly incorporates uncertainty estimation from the learned dynamics model.
The problem aims at finding a control trajectory $\mathbf{\bar{\uu}}$ such that the resulting trajectory $\mathbf{\bar{\x}}$ satisfies the planning requirements with high confidence.

\section{Methodology}
\label{sec:methodology}

In this work, we discretize the continuous dynamic system into a discrete-time formulation. Each control action corresponds to a parameterized push skill that is executed over a fixed duration, and the system is observed only at the terminal state. This design choice allows us to work with skill abstractions and is consistent with practical sensing conditions, where object poses are often observable only after the push concludes (e.g., due to vision occlusion). The system dynamics becomes:
\begin{equation}
    \Delta{\x_{n}} = f^*(\uu_{n})
\end{equation}

We represent each object as a two-dimensional oriented bounding box (OBB) and the planar push control is parameterized by three variables, as shown in \autoref{fig:foward_model}:
\begin{equation}
    \uu = (s, o, d)
\end{equation}
where $s \in \{1,\dots,4\}$ selects one of the box’s sides, $o$ is the lateral offset along that side from the center, and $d$ is the total push distance. The end-effector follows a straight-line fixed-duration ($\tau = 2\ \mathrm{s}$) sinusoidal velocity profile:
\begin{equation}
    v(t) = \frac{d}{\tau}\,\left[
    \sin\left(2\pi\,\frac{t}{\tau} - \frac{\pi}{2}\right) + 1
    \right]
\end{equation}

The system's state $x$ is defined as the object's \SETwo state $T$.
The effect of the push $f^*(\uu)$ is therefore defined as the \SETwo transformation between the object’s initial and final poses. Unlike Euclidean states with additive updates, evolution on transformation is expressed through multiplication:
\begin{equation}
\begin{gathered}
    T_{n+1} = T_n f^*(\uu_n) = T_n \Delta{T_n},
    \\
    f^*(\uu_n) = \Delta{T_n} = {T_n}^{-1} T_{n+1}
\end{gathered}
\end{equation}

\subsection{Residual Physics}
\label{subsec:residual}

In this section, we introduce the model $f(\uu)$, which will be trained to approximate the unknown true dynamics $f^*(\uu)$. To effectively predict in a low-data setting, we adopt the approach of learning residual physics, which integrates a physics-based model with a neural network~\cite{residual-pushing, tossingbot}. Rather than replacing the physics-based model, the neural network is tasked with learning the residual error, i.e. deviations from the idealized model output to the real observations. This preserves the physical plausibility while allowing the learned component to correct and improve overall accuracy.

\begin{figure}[ht!]
    \centering
    \includegraphics[width=\linewidth]{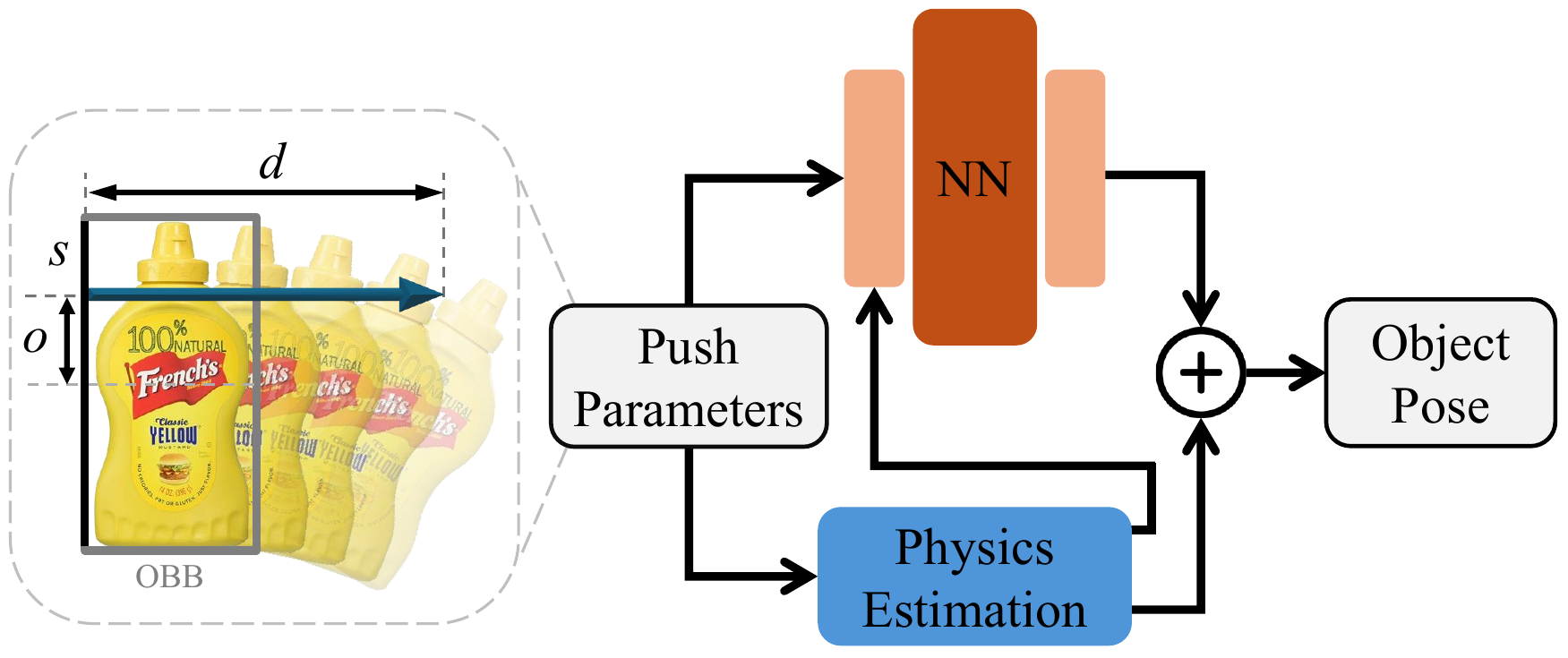}
    \caption{Neural network (NN) with residual physics architecture. The network takes both the control parameters and the output of the physics model to predict residuals, which are then added to the physics-based output to produce the final prediction.}
    \label{fig:foward_model}
\vspace{-2ex}
\end{figure}

For the analytical model, we follow the dynamic model proposed in~\cite{physics_equation} to predict object behavior given pusher motion. In this modeling, the object is treated as a rigid rectangle pushed under quasi-static assumption, with frictional forces obeying Coulomb’s law. The model further requires knowledge of the object’s shape and the ratio of frictional moment to frictional force $c_{ratio}$. However, we do not assume having access to these exact parameters. Thus, the analytic prediction serves only as a coarse estimation on how the object will move. Furthermore, to keep the model compatible with our neural network components and enable efficient batch operations, additional simplifications are applied. We set a fixed $c_{ratio} = 0.0187$ as recommended in~\cite{physics_equation}. Also, we assume the contact point is fixed and the force is perpendicular to the point, maintaining perfect sticking contact throughout the push. Please refer to our source code for more details.

As illustrated in \autoref{fig:foward_model}, our neural network takes both the skill parameters and the output of the physics equation as input. This design enables the network to reason about both the nominal dynamics and the data-driven corrections required to account for object-specific and contact-specific variations. The network finally outputs the residual, and the model combines it with the physics equation output to provide the final estimate.
We train the combined model by minimizing the \SETwo distance between prediction and observation, defined in Open Motion Planning Library (OMPL)~\cite{ompl}:
\begin{equation}
\label{eq:se2_dist}
\begin{gathered}
    \mathcal{D} (T_i, T_j) = 
    \mathcal{D}_{\R2} (p_i, p_j) + w_o \cdot \mathcal{D}_{\SOTwo} (r_i, r_j)
\end{gathered} 
\end{equation}
where $p \in \R2$ and $r \in \SOTwo$ are the positional and rotational component of the \SETwo state $T$. $\mathcal{D}_{\R2} (\cdot, \cdot)$ computes standard Euclidean norm between positions, while $\mathcal{D}_{\SOTwo} (\cdot, \cdot)$ measures the geodesic angular difference between two \SOTwo angles.
The scalar $w_o$ weights the rotational component. In this work, we set $w_o = 0.2$ to bias toward positional accuracy.

\subsection{Uncertainty Quantification}
\label{subsec:uncertainty}

\begin{figure}[ht!]
    \centering
    \includegraphics[width=0.9\linewidth]{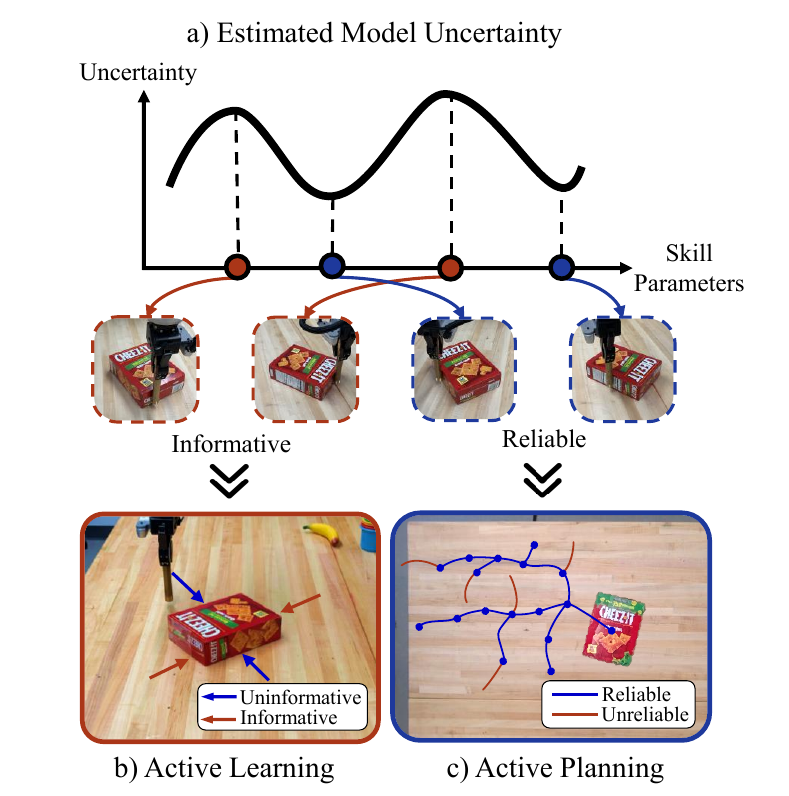}
    \caption{a) \method quantifies the model uncertainty of the learned model~(\autoref{subsec:uncertainty}).
    b) During the learning phase, \textsc{activePusher} chooses the most \textit{informative} push to apply to increase the learning efficiency~(\autoref{subsec:active_learning}).
    c) During planning the most \textit{reliable} pushes are chosen to maximize the task success rate~(\autoref{subsec:active_planning}).}
    \label{fig:uncertainty}
    \vspace{-4ex}
\end{figure}

Traditionally, neural network-based dynamics models produce only point estimates of action outcomes, lacking a measure of their prediction uncertainty. By explicitly quantifying the epistemic uncertainty in the learned model, \method enables both informative data acquisition and uncertainty-aware robust planning, as illustrated in \autoref{fig:uncertainty}. During learning, this uncertainty guides active data collection by prioritizing the most informative samples in under-explored regions, thereby improving data efficiency (\autoref{subsec:active_learning}). At execution time, the planner leverages this uncertainty to select reliable actions from well-explored regions of the action space, resulting in more robust planning (\autoref{subsec:active_planning}).

\method estimates model uncertainty by leveraging the correspondence between neural networks and Gaussian Processes (GP) with the Neural Tangent Kernel (\NTK)~\cite{ntk}. 
Under this view, the epistemic uncertainty of the learned dynamics model can be approximated through the predictive covariance of a \GP with an \NTK kernel.

Given a fully-connected neural network $f_{\theta}(u)$ with infinite width, parameterized by weights $\theta$, its \NTK stays constant during training and is defined as 
\begin{equation}
\label{eq:ntk}
    k_{\NTK}(u, u') \;=\;
    \langle\;
    \nabla_{\theta}f_{\theta}(u),\,
    \nabla_{\theta}f_{\theta}(u')
    \;\rangle\,
\end{equation}
where $\nabla_{\theta}f_{\theta}(u)$ denotes the gradient of the network output with respect to the parameters $\theta$, $u$ and $u'$ are two different inputs, and $\langle \cdot, \cdot\rangle$ is the inner product of two gradient vectors.

After training with gradient descent to convergence, the predictive distribution of an infinite-width neural network \(f_{\theta}(u)\) is equivalent to that of Gaussian process regression with the \NTK as its kernel~\cite{ntk}.
Therefore, its predictive distribution, capturing model epistemic uncertainty, can be approximated by a \GP with the \NTK:
\begin{equation}
\label{eq:gp}
    \mathcal{GP} (0, k_{\NTK}(u, u')).
\end{equation}

\NTK encodes a notion of similarity between two inputs from the perspective of the neural network. The gradient $\nabla_{\theta}f_{\theta}(u)$ reflects how sensitive the network's prediction $f_{\theta}(u)$ is to infinitesimal perturbations in parameter $\theta$. 
Intuitively, two inputs $u$ and $u'$ are considered similar if their gradients are aligned. In this case, model parameter updates that improve the prediction at $u$ will also tend to improve the prediction at $u'$. 
This similarity is directly relevant to epistemic uncertainty, as the model is less uncertain in regions where multiple inputs share aligned gradients. This notion of similarity is precisely captured by the \NTK, where gradient similarity is computed.

For finite‐width networks, the kernel will change during the training process. But in practice, \NTK after training convergence (empirical \NTK) still provides accurate model uncertainty estimates~\cite{ntk, lookahead_active}.
Assume we train neural network model $f(\uu)$ with training data $\set_{train}$ and their corresponding observed labels $\mathcal{L}_{train}$. Here, a set of data $\set$ is a collection of skill action $u$. As mentioned, we can model $f$ as a \GP with $k_{\NTK}$.
In addition, we consider a data pool $\set_\text{pool}$, which is the set of candidate actions whose labels have not yet been observed.
Conditioned on the observed $\set_{train}$ and $\mathcal{L}_{train}$, the resulting posterior predictive covariance enables estimation of model uncertainty on unobserved data points within $\set_\text{pool}$. 
Formally, given a pre-defined inherent data noise $\sigma_d$, the posterior predictive covariance of the \GP over an unlabeled data pool $\set_\text{pool}$ is:
\begin{equation}
\label{eq:epistemic}
\begin{aligned}
\mathrm{Cov}(\set_\text{pool})
&= K_{\NTK}(\set_\text{pool}, \set_\text{pool}) \\
&- K_{\NTK}(\set_\text{pool}, \set_\text{train})
     K_t^{-1}
     K_{\NTK}(\set_\text{train}, \set_\text{pool}), \\
\text{where } K_t
&= K_{\NTK}(\set_\text{train}, \set_\text{train}) + \sigma_d^{2} I .
\end{aligned}
\end{equation}
where $K_{\NTK}(\cdot, \cdot)$ is the Gram matrix of pairwise \NTK values between two vector inputs, capturing similarity of two sets of the data.
We set $\sigma_d = 0.005$ in our experiments. Since $\sigma_d$ is fixed, the posterior covariance of a \GP (\autoref{eq:epistemic}) primarily reflects epistemic uncertainty arising from limited training data.
By isolating the diagonal terms of the posterior covariance matrix, we obtain per-sample model uncertainty estimates for $\set_{pool}$:
\begin{equation}
\label{eq:var_epistemic}
    \mathrm{Var}(\set_\text{pool}) = \mathrm{Diag}(\mathrm{Cov}(\set_\text{pool}))
\end{equation}

%
\subsection{Active Learning}
\label{subsec:active_learning}

\begin{algorithm}
    \caption{\texttt{Active Learning}}
    \label{alg:active-learning}
    \SetNoFillComment
    \KwIn{Kernel $k$, training round $N$, batch size $B$, initial training set $\set_{train}$, pool set $\set_{pool}$}
    \For{$i \gets 1$ \textbf{to} $N$}{
        $\set_{sel} \gets \texttt{Acquire}(k, \set_{train}, \set_{pool}, B)$\;
        Acquire labels of $\set_{sel}$\;
        $\set_{train} \gets \set_{train} \cup \set_{sel}$\;
        $\set_{pool} \gets \set_{pool} \setminus \set_{sel}$\;
        Train model on $\set_{train}$ with corresponding labels and update kernel $k$\;
    }
\end{algorithm}

Rather than passively training on a randomly collected dataset, active learning tries to iteratively query the most informative batch of samples to improve model performance with fewer labels~\cite{bald, bait}. The general active learning process in a kernel setting is defined in \autoref{alg:active-learning}. In each of the $N$ training rounds, we perform uncertainty estimation over all unlabeled data in $\set_\text{pool}$ and select $\set_\text{sel}$ set with $B$ informative samples. After querying their labels and moving them into the training set $\set_\text{train}$, we retrain the model with the expanded $\set_\text{train}$ and proceed to the next round.

Given predictive uncertainty estimation, one can apply different data acquisition strategies. In this work, we adopt the \BAIT algorithm~\cite{bait} with Fisher information to actively select the most informative action batch to collect. 

Intuitively, the Fisher information encodes how sensitively the model’s predictions respond to changes in its parameters. Covering the Fisher space ensures that the selected batch spans the directions along which the model can still learn the most. \BAIT aims to select $B$ samples whose combined per-sample Fisher embeddings best approximate (in Frobenius norm) the global Fisher information, yielding a representative batch that jointly captures both model uncertainty and diversity. 
Specifically, it seeks to minimize the trace of the inverse Fisher information matrix of the selected batch (i.e., the model uncertainty after selecting a batch), pre-multiplied by the Fisher information of the entire unlabeled pool:
\begin{equation}
\label{eq:fisher_objective}
\set_{\text{sel}} = 
\mathop{\arg\min}_{\set_{\text{sel}} \subseteq \set_{\text{pool}}}
\operatorname{tr}\left(
\left(
\sum_{u \in \set_{\text{sel}}} I(u; \theta)
\right)^{-1}
\left(
\sum_{u \in \set_{\text{pool}}} I(u; \theta)
\right)
\right)
\end{equation}
where $I(u; \theta)$ is the Fisher information matrix associated with model parameter $\theta$ at input $u$. In Gaussian regression, the Fisher information matrix of a set $\set$ reduces to the outer product of the output gradients:
\begin{equation}
\label{eq:fisher_gaussian}
\sum_{u \in \set} I(u; \theta)
=
\nabla_{\theta} f_{\theta}(\set)^\top\, \nabla_{\theta} f_{\theta}(\set)
\end{equation}

Let $k[\set]$ denote the kernel $k$ conditioned on $\set$, as shown in~\cite{activeregression}, one can prove that, with \NTK kernel:
\begin{equation}
\label{eq:kernel_bait}
\begin{gathered}
\sum_{u \in \set_{pool}} 
k[\set_{sel} ](u, u)
= 
c \operatorname{tr} \left(
G_{\set_{sel}}^{-1}
\;
G_{\set_{pool}}
\right),
\\
\text{where } G_{\set} := \nabla_{\theta} f_{\theta}(\set)^\top\, \nabla_{\theta} f_{\theta}(\set)
\end{gathered}
\end{equation}
Combined with \autoref{eq:fisher_gaussian}, we show that optimizing \autoref{eq:kernel_bait} is equivalent to optimizing the Fisher objective \autoref{eq:fisher_objective} in \BAIT.

Different from the original \BAIT, we made the following changes. First, considering that our neural network is relatively small, we use the full gradient \NTK instead of the last-layer gradient as in the original design. Second, rather than computing the Fisher information of merely the selected set $\set_{sel}$ and pool set $\set_{pool}$, we expand \autoref{eq:kernel_bait} by also considering the current training set $\set_{train}$. These changes better exploit the full representational capacity of the network and ensure that acquisition decisions account for both the unlabeled pool and the knowledge already contained in the training set. In summary, our acquisition strategy is as follows:
\begin{equation}
\label{eq:uncertainty}
\begin{gathered}
\texttt{Acquire}(k, \set_{train}, \set_{pool}, B) = 
\\
\mathop{\arg\min}_{\set_{sel} \subseteq \set_{pool}} \,
\underset{u \in \set_{train} \cup \set_{pool}}{\sum} \,
k_{\NTK}[\set_{train} \cup \set_{sel}](u, u)
\end{gathered}
\end{equation}

Optimizing such a Fisher objective is intractable given the many potential different combinations for $\set_{sel}$. To address this, the same greedy forward–backward selection algorithms, proposed in~\cite{bait}, are adopted to greedily select $\set_{sel}$.

\subsection{Active Planning}
\label{subsec:active_planning}

We formulate nonprehensile pushing as a kinodynamic planning problem in the object’s \SETwo state space. In this formulation, each parameterized push action $\uu$ becomes a discrete control that drives the object’s state through forward simulation of the learned dynamics. We use an asymptotically near optimal kinodynamic planner, specifically \sst \cite{liSST2016}, to explore the object’s state space directly. \sst incrementally expands a tree of dynamically feasible trajectories by forward-propagating sampled controls and pruning redundant nodes to maintain sparsity. Optionally, given a optimization objective, it progressively improves trajectory cost while steering the search toward the goal region.

In the absence of model error, control sequences found by \sst succeed by design. In practice, however, accumulated prediction errors in the learned dynamics  can lead to execution failures. 
To improve robustness, we integrate epistemic uncertainty estimates into the control sampling step. 

\begin{algorithm}
\caption{\texttt{Active Sampling}}
\label{alg:active_sampling}
\KwIn{Kernel $k$, training set $\set_{\text{train}}$, batch size $b$}

$ \set_U \gets \texttt{random\_sampling}(b) $\;

$ \mathrm{Var} \gets \texttt{query\_uncertainty}(k, \set_{\text{train}}, \set_U) \; (\autoref{eq:var_epistemic})$\;

$u \gets \mathrm{argmin}_{u \in \set_U} \mathrm{Var}[u] $\;

\Return $u$

\end{algorithm}
Specifically, as summarized in \autoref{alg:active_sampling}, at each planning step, a batch of candidate pushing controls of size $b$ is first sampled. We then evaluate the epistemic uncertainty of the learned model for these controls (\autoref{eq:var_epistemic}). Instead of selecting randomly, we choose the control with the lowest predicted uncertainty. This strategy biases control sampling in planning toward more reliable actions, which is based on the intuition that the model is more accurate in well-explored regions of the skill space.

Because the biased active sampler still samples from all the control space (i.e., every action has non-zero probability of being sampled), the planner retains the probabilistic completeness and asymptotic optimality guarantees of \sst. In effect, uncertainty estimates guide the planner to prioritize reliable regions of the skill space while maintaining theoretical coverage of the entire domain.

\section{Experiments}
\label{sec:experiments}

In this section, we provide details of our models, training procedure and the experiments conducted in both simulation and real world.
Our experiment setup includes a 6-\dof \textsc{ur10} robot and multiple objects from the YCB dataset \cite{ycb}, with different geometric shapes and physical characteristics, to test the robustness of our approach across a range of physical characteristics. 
All the learning and planning experiments are run on a workstation with an NVIDIA RTX 4070 Ti Super GPU and 32GB of RAM.

In real world, object pose is estimated with FoundationPose~\cite{foundationpose} using an Intel RealSense Depth Camera D435 mounted on the end effector.
Additionally, for real-world data collection, we implement a simple reset algorithm that automatically pushes the object back toward the center of the robot’s high‑manipulability workspace whenever it drifts outside that region. This mechanism enables autonomous, continuous learning without any human intervention.
To execute the push parameters with the robot, we use a global redundancy resolution method \cite{zhong2024grr} as inverse kinematics (IK) solutions, which guarantees valid and smooth joint trajectories within a certain workspace region.

\begin{figure}[h]
    \centering
    \includegraphics[width=0.95\linewidth]{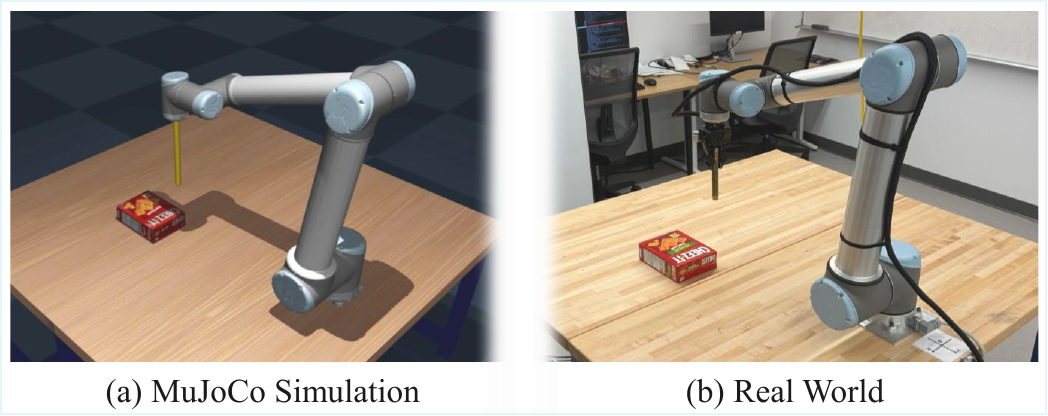}
    \caption{Experiment Setup}
    \label{fig:setup}
\vspace{-2ex}
\end{figure}

We perform our simulation experiments with the MuJoCo Simulator \cite{mujocoplayground}, which allows parallel execution of multiple environments, enabling efficient large-scale testing. The simulated scene in the simulation environment replicates our real-world setup, as shown in \autoref{fig:setup}.


We evaluate the proposed method in two settings: (i) Skill Learning, where we measure active learning performance, and (ii) long-horizon Kinodynamic Planning, where we assess the effectiveness of both active learning and planning.


\subsection{Skill Learning}
\label{subsec:exp_skill}
As discussed in \autoref{subsec:active_learning}, active learning can run continuously to collect training data and update models. However, for the purpose of repetitive evaluation and fair comparison across methods, we adopt a standard pool-based evaluation setup~\mbox{\cite{bald, bait}}. In simulation, we first construct a candidate pool of 9,000 push actions for each object, while in the real world, each object has a pool of 1,000 actions. During training, at each acquisition stage, a batch of 10 samples $\set_{sel}$ is selected from the pool to expand the training set and update the model. This process is repeated for 10 stages until 100 data are collected. We collect independent test sets of 1,000 samples for each object in either simulation or real world. 

Four objects in simulation (Banana, Mug, Cracker Box and Mustard Bottle) and two objects in the real world (Cracker Box and Mustard Bottle) are used, and we evaluate the following five methods:

\begin{itemize}
\item \textit{Pure Physics:} The analytical dynamics model adapted from ~\cite{physics_equation}.
\item \textit{MLP Random:} A fully connected multi-layer perceptron (MLP) consisting of five hidden layers with sizes [32, 64, 64, 32, 32] trained with random push actions.
\item \textit{Residual Random:} The hybrid model as described in \autoref{subsec:residual}. The physics is the same as \textit{Pure Physics}. Meanwhile, the neural network architecture and data collection method are the same as \textit{MLP Random}.
\item \textit{MLP BAIT:} The same MLP architecture as \textit{MLP Random}, but trained via our \NTK‐driven active learning pipeline to collect informative data.
\item \textit{Residual BAIT:} The hybrid residual model as in \textit{Residual Random}, but trained with active learning pipeline. 
\end{itemize}
All models are trained using a learning batch size of 20 for 1000 epochs until convergence. 
\begin{figure}
    \centering
    \includegraphics[width=1.0\linewidth]{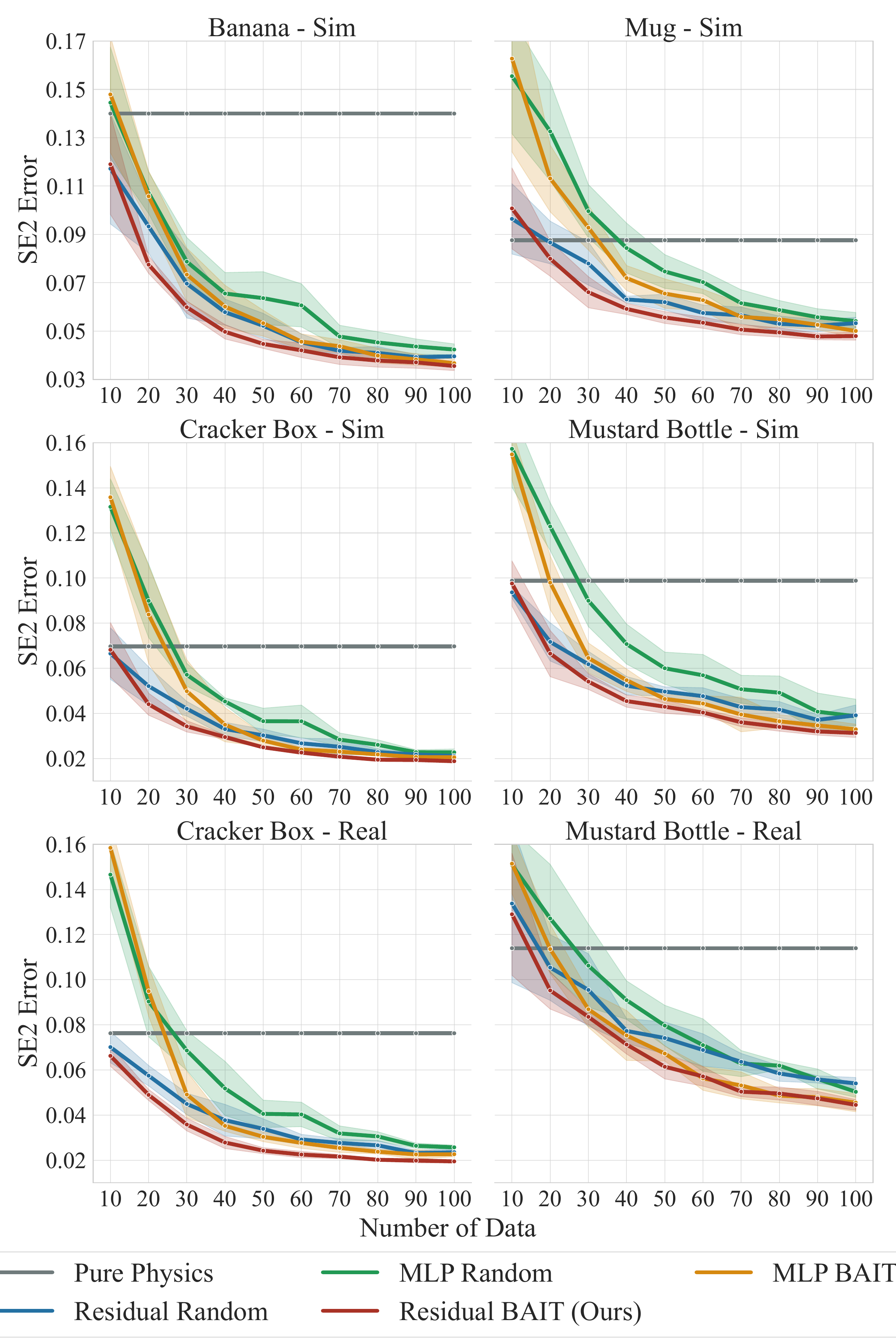}
    \caption{Skill Learning results show SE2 prediction error for 2 real objects, and 4 simulated objects from~\cite{ycb}: Banana, Mug, Cracker Box and Mustard Bottle. The active learning methods outperform random data collection, and models with residual physics perform better in low-data regime.}
    \vspace{-4ex}
    \label{fig:active_learning}
\end{figure}
Each training is repeated 5 times and the summarized results for prediction \SETwo error~(\autoref{eq:se2_dist}) on the test set are shown in~\autoref{fig:active_learning}. The models with residual physics show a clear improvement when data is limited. Additionally, the active learning approach consistently outperforms random sampling across all objects, either by reaching the same level of accuracy with fewer training samples or by achieving higher accuracy given the same amount of data. 
By jointly leveraging active learning and analytical physics, our method \textit{Residual BAIT} can achieve comparable accuracy to the final performance of baseline \textit{MLP Random} while requiring only 55\% of the training data on average.
In practice, active learning can be run continuously, with data collection terminated once validation performance converges, or according to a task-specific criterion~\cite{kumar2024practice}.

\subsection{Kinodynamic Planning}
\label{subsec:exp_planning}
In this experiment, we demonstrate the performance of learned models integrated with a kinodynamic planner for two downstream tasks. The planning was conducted in the object’s state space, defined as $\X = \SETwo$, with a control space $\mathcal{U} \subset \R{3} $ corresponding to the pushing parameters described in \autoref{sec:methodology}. The valid state space $\Xfree$ is constrained to the table surface and are free of collisions with obstacles.

\begin{figure}[h]
    \centering
    \includegraphics[width=0.9\linewidth]{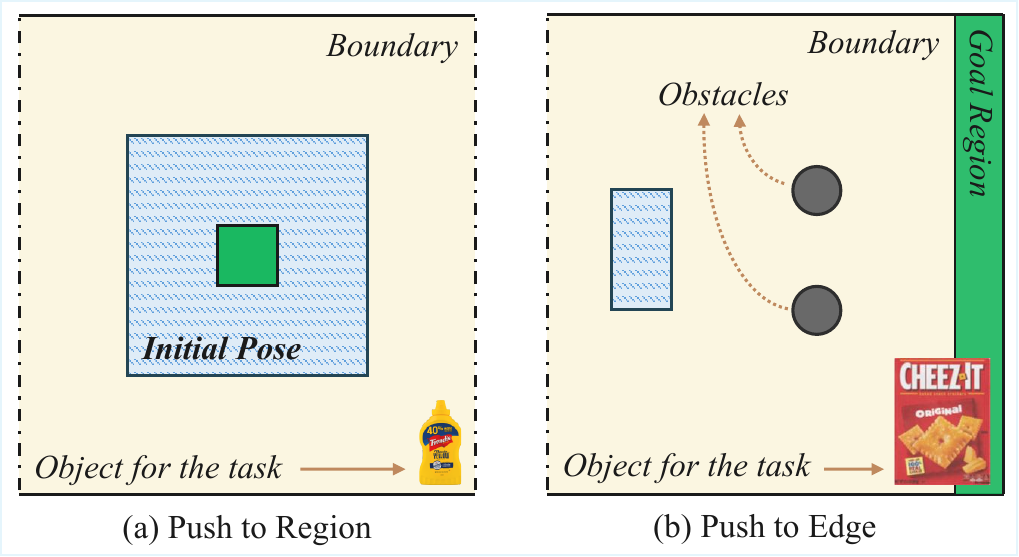}
    \caption{2D schematic of the planning tasks}
    \label{fig:planning_tasks}
    \vspace{-4ex}
\end{figure}

The two task environments are shown in \autoref{fig:planning_tasks}. In the simpler "Push to Region" task, a mustard bottle is initialized randomly and needs to be pushed to a $0.1 \text{m} \times 0.1 \text{m}$ region at the table center. The more challenging "Push to Edge" requires pushing a non-graspable cracker box toward the table edge to enable a feasible pick-up, with two cylinder obstacles (radius $0.04 \text{m}$) present. The goal is defined geometrically: the object’s center of mass remains on the table while at least one corner extends beyond $0.03 \text{m}$ from the boundary.

We use \sst \cite{liSST2016} in \textsc{ompl} \cite{ompl} with a path-length optimization objective to plan for both tasks. In simulation, we solve 100 randomly generated problems for each task and repeat the planning 5 times. For the real world experiment, we solve 20 randomly generated problems once. For all the problems, we compare the default uniform sampling method (Regular Planning) with our proposed active sampling method (Active Planning). To also assess the impact of model prediction accuracy on planning outcomes, we evaluate baseline model \textit{MLP Random} and proposed model \textit{Residual BAIT}, described in the previous section \autoref{subsec:exp_skill}, with different data sizes.

\begin{figure}
  \centering
  \includegraphics[width=1.0\linewidth]{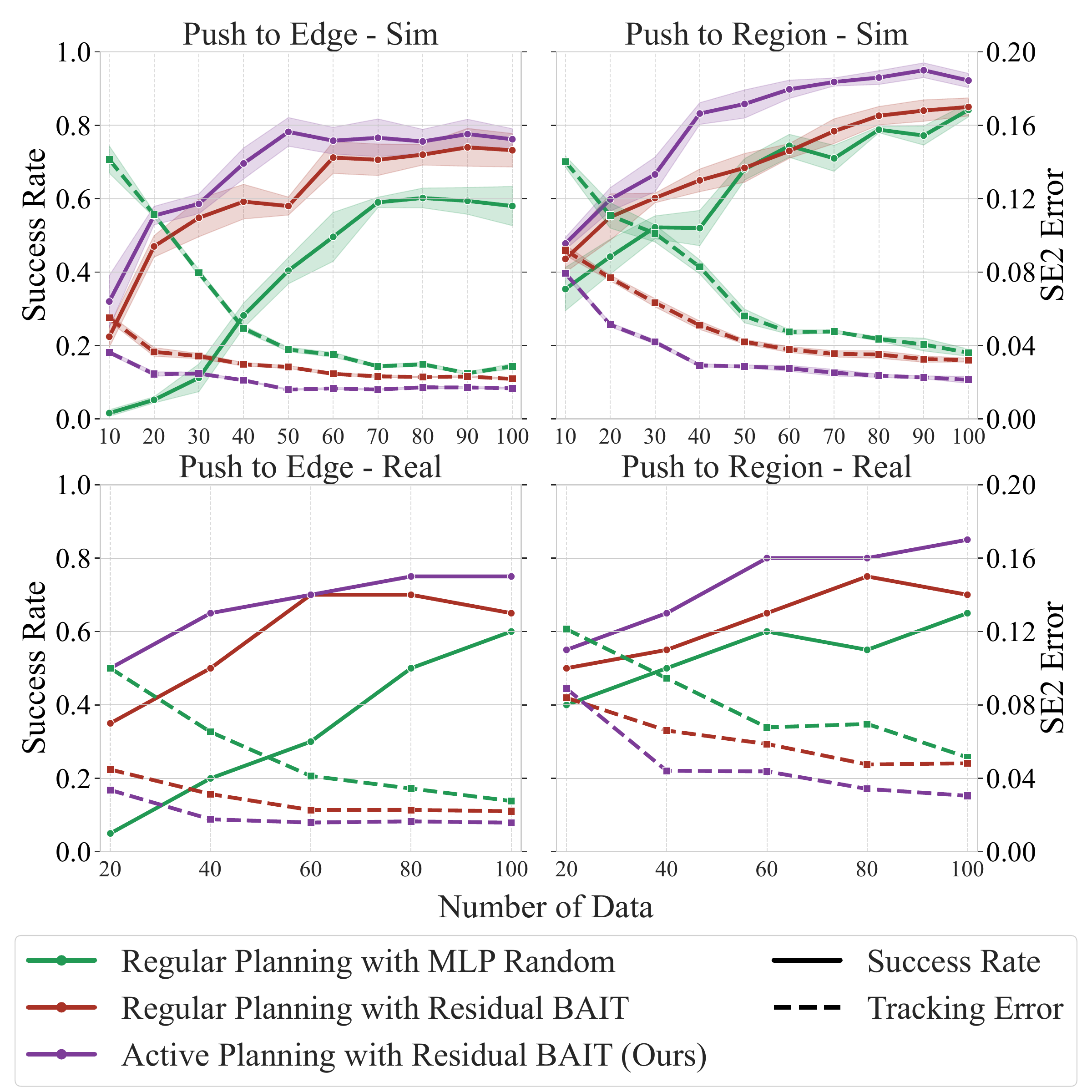}
  \caption{Kinodynamic Planning results show that more accurate dynamics models lead to higher task success rates. Additionally, active planning tends to select actions with lower execution error and further improves task success.}
  \label{fig:real_planning}
  \vspace{-2ex}
\end{figure}

All generated plans are executed in an open-loop manner. Performance is evaluated by task success rate and path-tracking error. A task is considered successful if the goal is reached while maintaining validity, and the path-tracking error is defined as the average $\SETwo$ deviation along the trajectory. 
The results are summarized in~\autoref{fig:real_planning}.

The results of the experiment show that the planning performance is closely tied to the model prediction accuracy as expected. More accurate models generally produce plans with higher task success rates and lower execution errors. In addition, incorporating active action sampling further improves performance.
By steering the planner toward actions in which the model is more confident, active planning tends to select actions with potentially lower execution error. 
As a result, active planning consistently outperforms regular planning in both success rate and tracking accuracy.

We note that active planning produces slightly longer paths, as the planner favors sampling actions in lower-uncertainty regions.
In our experiments, this results in an increase of approximately $9–13\%$ in average path length, while improving success rate and execution accuracy.
The additional computational overhead is negligible in practice. Estimating the model uncertainty for 10,000 candidate actions takes only $8.1$ millisecond on average in a batched implementation with neural-network gradient features.
%

\begin{figure}
  \centering
  \includegraphics[width=0.9\linewidth]{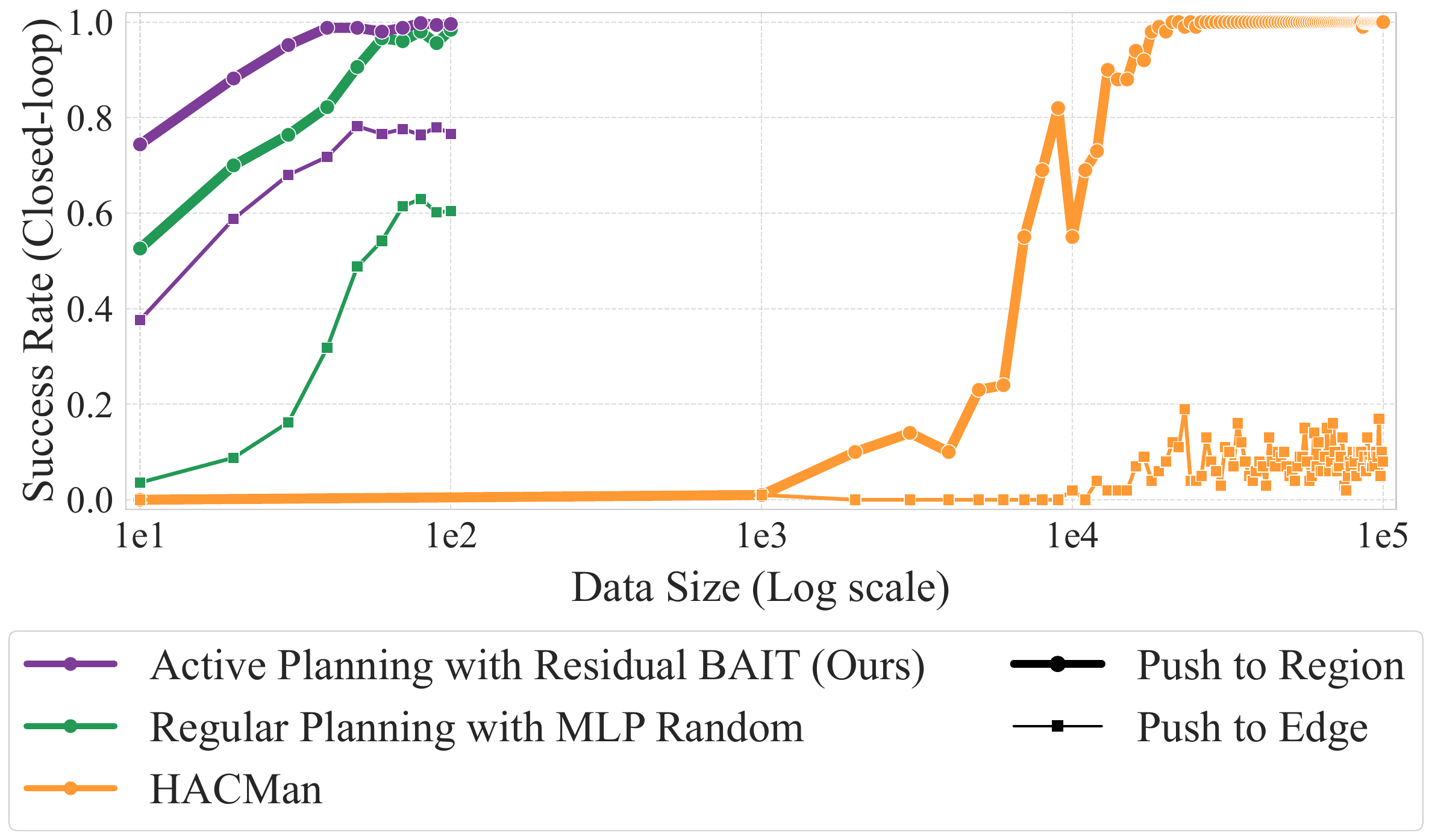}
  \caption{Closed-loop Execution results show our method outperforms HACMan, especially in Push to Edge task, while requiring far less data and transferring effectively to complex environments with novel obstacles.}
  \label{fig:closed_loop}
  \vspace{-4ex}
\end{figure}

\subsection{Closed-loop Execution}

All previous experiments are conducted in an open-loop setting. In practice, replanning given current state can improve robustness. To examine this, we repeat the same tasks in a closed-loop manner. We define a \emph{recoverable failure} as an execution that does not reach the goal but remains in the valid state space. The system can replan within 1 second and continue execution. In contrast, \emph{irrecoverable failures} occur when the object collides with obstacles, falls off the table, or exceeds the planning horizon of 10 control steps.
We additionally compare against HACMan~\cite{zhou2023hacman}, a state-of-the-art RL method that executes closed-loop, with discrete actions learned entirely from interactions. Results are in \autoref{fig:closed_loop}.

On the \textit{Push to Region} task, \method can reach $100\%$ success rate with replanning. HACMan can also reach $100\%$ success but requires orders of magnitude more data. On the more challenging \textit{Push to Edge} task, HACMan struggles with distribution shift: obstacles are randomized during training but differ in placement during evaluation. In contrast, our method requires far less data and transfers effectively to novel obstacles and goal constraints. 

Additionally, sim-to-real transfer remains a challenge for RL-based approaches. As reported in~\cite{zhou2023hacman}, HACMan achieves only about $70\%$ success in simple planar push tasks without obstacles. In contrast, our method can be trained directly and efficiently with limited real-world data, achieving robust performance without additional adaptation even with obstacles.

Also, closed-loop replanning alone does not eliminate all failures. In the \textit{Push to Edge} task, an inappropriate initial plan can drive the object into irrecoverable states. This emphasizes the importance of generating robust initial plans.

\section{Conclusion and Future Work}
\label{sec:conclusion}

In this paper, we present \method, a framework that integrates residual physics, active learning, and active kinodynamic planning.
By modeling epistemic uncertainty using \NTK, our method actively gathers informative data with \BAIT strategy, and biases planning toward reliable actions.
Our experiments demonstrate that \method achieves higher prediction accuracy and planning success with fewer interactions compared to baselines. 
This integration of learning and planning offers a promising path toward data-efficient and reliable nonprehensile manipulation.

Our current formulation relies on several simplifying assumptions to enable analytical modeling and a relatively small neural network.
The major simplification is to approximate rigid objects as oriented bounding boxes, and assume sticking contact during pushing. 
These assumptions allow efficient learning and uncertainty estimation but limit the applicability of the framework to planar quasi-static nonprehensile manipulation.
Importantly, the core contribution of this work is uncertainty estimation for both active learning and planning, which is largely independent of the specific problem modeling, and can be integrated with more sophisticated modeling approaches.
Future work will broaden the scope beyond simple planar pushing toward richer contact dynamics, diverse geometries, and \SEThree nonprehensile skills. 
Another direction is to jointly model epistemic and aleatoric uncertainty to better capture action-dependent noise.


\bibliographystyle{IEEEtran_ShortURL}
\bibliography{reference}

@ARTICLE{tossingbot,
  author={Zeng, Andy and Song, Shuran and Lee, Johnny and Rodriguez, Alberto and Funkhouser, Thomas},
  journal={IEEE Transactions on Robotics}, 
  title={TossingBot: Learning to Throw Arbitrary Objects With Residual Physics}, 
  year={2020},
  volume={36},
  number={4},
  pages={1307-1319},
  url={https://ieeexplore.ieee.org/document/9104757}
}

@INPROCEEDINGS{residual-pushing,
  author={Ajay, Anurag and Wu, Jiajun and Fazeli, Nima and Bauza, Maria and Kaelbling, Leslie P. and Tenenbaum, Joshua B. and Rodriguez, Alberto},
  booktitle= {IEEE/RSJ International Conference on Intelligent Robots and Systems (IROS)}, 
  title={Augmenting Physical Simulators with Stochastic Neural Networks: Case Study of Planar Pushing and Bouncing}, 
  year={2018},
  volume={},
  number={},
  pages={3066-3073},
  url={https://ieeexplore.ieee.org/document/8593995}
}

@inproceedings{poking,
 author = {Agrawal, Pulkit and Nair, Ashvin V and Abbeel, Pieter and Malik, Jitendra and Levine, Sergey},
 booktitle = {Advances in Neural Information Processing Systems},
 pages = {},
 publisher = {Curran Associates, Inc.},
 title = {Learning to Poke by Poking: Experiential Learning of Intuitive Physics},
 url = {https://proceedings.neurips.cc/paper_files/paper/2016/file/c203d8a151612acf12457e4d67635a95-Paper.pdf},
 volume = {29},
 year = {2016}
}

@ARTICLE{ycb,
  author={Calli, Berk and Walsman, Aaron and Singh, Arjun and Srinivasa, Siddhartha and Abbeel, Pieter and Dollar, Aaron M.},
  journal={IEEE Robotics \& Automation Magazine}, 
  title={Benchmarking in Manipulation Research: Using the Yale-CMU-Berkeley Object and Model Set}, 
  year={2015},
  volume={22},
  number={3},
  pages={36-52},
  doi={10.1109/MRA.2015.2448951},
  url={https://ieeexplore.ieee.org/document/7254318}
}

@inproceedings{bald,
  title={Deep Bayesian active learning with image data},
  author={Gal, Y and Islam, R and Ghahramani, Z},
  booktitle={Proceedings of the 34th International Conference on Machine Learning },
  volume={70},
  year={2017},
  organization={PMLR},
  url={https://proceedings.mlr.press/v70/gal17a.html}
}

@article{bait,
  title={Gone fishing: Neural active learning with fisher embeddings},
  author={Ash, Jordan and Goel, Surbhi and Krishnamurthy, Akshay and Kakade, Sham},
  journal={Advances in Neural Information Processing Systems},
  volume={34},
  pages={8927--8939},
  year={2021},
  url = {https://arxiv.org/abs/2106.09675},
}

@article{ntk,
  title={Neural tangent kernel: Convergence and generalization in neural networks},
  author={Jacot, Arthur and Gabriel, Franck and Hongler, Cl{\'e}ment},
  journal={Advances in neural information processing systems},
  volume={31},
  year={2018},
  url={https://arxiv.org/abs/1806.07572},
}

@article{activeregression,
  title={A framework and benchmark for deep batch active learning for regression},
  author={Holzm{\"u}ller, David and Zaverkin, Viktor and K{\"a}stner, Johannes and Steinwart, Ingo},
  journal={Journal of Machine Learning Research},
  volume={24},
  number={164},
  pages={1--81},
  year={2023},
  url = {https://dl.acm.org/doi/abs/10.5555/3648699.3648863}
}

@article{ltamp,
  author = {Zi Wang and Caelan Reed Garrett and Leslie Pack Kaelbling and Tomás Lozano-Pérez},
  title ={Learning compositional models of robot skills for task and motion planning},
  journal = {The International Journal of Robotics Research},
  volume = {40},
  number = {6-7},
  pages = {866-894},
  year = {2021},
  URL = {https://doi.org/10.1177/02783649211004615},
}

@article{activeleraningcontrol,
  title = {Active learning in robotics: A review of control principles},
  journal = {Mechatronics},
  volume = {77},
  pages = {102576},
  year = {2021},
  issn = {0957-4158},
  url = {https://www.sciencedirect.com/science/article/pii/S0957415821000659},
  author = {Annalisa T. Taylor and Thomas A. Berrueta and Todd D. Murphey},
}

@INPROCEEDINGS{zhong2024grr,
  author={Zhong, Zhuoyun and Li, Zhi and Chamzas, Constantinos},
  booktitle={IEEE/RSJ International Conference on Intelligent Robots and Systems (IROS)}, 
  title={Expansion-GRR: Efficient Generation of Smooth Global Redundancy Resolution Roadmaps}, 
  year={2024},
  volume={},
  number={},
  pages={8854-8860},
  doi={10.1109/IROS58592.2024.10801917},
  url={https://ieeexplore.ieee.org/document/10801917},
}

@inproceedings{lagrassa2024task,
  title={Task-oriented active learning of model preconditions for inaccurate dynamics models},
  author={LaGrassa, Alex and Lee, Moonyoung and Kroemer, Oliver},
  booktitle={IEEE International Conference on Robotics and Automation (ICRA)},
  pages={16445--16445},
  year={2024},
  url={https://ieeexplore.ieee.org/abstract/document/10611488},
}

@inproceedings{ren2023kinodynamic,
  title={Kinodynamic rapidly-exploring random forest for rearrangement-based nonprehensile manipulation},
  author={Ren, Kejia and Chanrungmaneekul, Podshara and Kavraki, Lydia E and Hang, Kaiyu},
  booktitle={IEEE International Conference on Robotics and Automation (ICRA)},
  pages={8127--8133},
  year={2023},
  url={https://ieeexplore.ieee.org/abstract/document/10161560},
}

@inproceedings{bauza2018data,
  title={A data-efficient approach to precise and controlled pushing},
  author={Bauza, Maria and Hogan, Francois R and Rodriguez, Alberto},
  booktitle={Conference on Robot Learning},
  pages={336--345},
  year={2018},
  organization={PMLR},
  url={https://proceedings.mlr.press/v87/bauza18a.html},
}

@inproceedings{wang2024uno,
  title={UNO Push: Unified Nonprehensile Object Pushing via Non-Parametric Estimation and Model Predictive Control},
  author={Wang, Gaotian and Ren, Kejia and Hang, Kaiyu},
  booktitle={IEEE/RSJ International Conference on Intelligent Robots and Systems (IROS)},
  pages={9893--9900},
  year={2024},
  url={https://ieeexplore.ieee.org/document/10802843},
}

@article{orthey2024-review-sampling,
  author = {Orthey, Andreas and Chamzas, Constantinos and Kavraki, Lydia E.},
  title = {Sampling-Based Motion Planning: A Comparative Review},
  journal = {Annual Review of Control, Robotics, and Autonomous Systems},
  volume = {7},
  number = {1},
  pages = {285-310},
  year = {2024},
  url = {https://doi.org/10.1146/annurev-control-061623-094742},
  month = jul
}

@article{kumar2024practice,
  title={Practice Makes Perfect: Planning to Learn Skill Parameter Policies},
  author={Kumar, Nishanth and Silver, Tom and McClinton, Willie and Zhao, Linfeng and Proulx, Stephen and Lozano-P{\'e}rez, Tom{\'a}s and Kaelbling, Leslie Pack and Barry, Jennifer},
  journal={Planning},
  volume={1},
  pages={2},
  url={https://www.roboticsproceedings.org/rss20/p040.pdf},
}

@article{hogan2020reactive,
author = {Francois R Hogan and Alberto Rodriguez},
title ={Reactive planar non-prehensile manipulation with hybrid model predictive control},
journal = {The International Journal of Robotics Research},
volume = {39},
number = {7},
pages = {755-773},
year = {2020},
doi = {10.1177/0278364920913938},
URL = {https://doi.org/10.1177/0278364920913938},
}

@article{mason2018toward,
   author = "Mason, Matthew T.",
   title = "Toward Robotic Manipulation", 
   journal= "Annual Review of Control, Robotics, and Autonomous Systems",
   year = "2018",
   volume = "1",
   number = "Volume 1, 2018",
   pages = "1-28",
   url = "https://www.annualreviews.org/content/journals/10.1146/annurev-control-060117-104848",
   publisher = "Annual Reviews",
   issn = "2573-5144",
   type = "Journal Article"
  }

@INPROCEEDINGS{physics_equation,
  author={Lynch, K.M. and Maekawa, H. and Tanie, K.},
  booktitle={Proceedings of the IEEE/RSJ International Conference on Intelligent Robots and Systems}, 
  title={Manipulation And Active Sensing By Pushing Using Tactile Feedback}, 
  year={1992},
  volume={1},
  number={},
  pages={416-421},
  doi={10.1109/IROS.1992.587370},
  url={https://ieeexplore.ieee.org/abstract/document/587370},
}

@article{liSST2016,
author = {Yanbo Li and Zakary Littlefield and Kostas E. Bekris},
title ={Asymptotically optimal sampling-based kinodynamic planning},
journal = {The International Journal of Robotics Research},
volume = {35},
number = {5},
pages = {528-564},
year = {2016},
doi = {10.1177/0278364915614386},
URL = { https://doi.org/10.1177/0278364915614386},
eprint = { https://doi.org/10.1177/0278364915614386},
}

@article{ompl,
    Author = {Ioan A. {\c{S}}ucan and Mark Moll and Lydia E. Kavraki},
    Doi = {10.1109/MRA.2012.2205651},
    Journal = {{IEEE} Robotics \& Automation Magazine},
    Month = {December},
    Number = {4},
    Pages = {72--82},
    Title = {The {O}pen {M}otion {P}lanning {L}ibrary},
    Note = {\url{https://ompl.kavrakilab.org}},
    Volume = {19},
    Year = {2012}
}

@article{lookahead_active,
  title={Making look-ahead active learning strategies feasible with neural tangent kernels},
  author={Mohamadi, Mohamad Amin and Bae, Wonho and Sutherland, Danica J},
  journal={Advances in Neural Information Processing Systems},
  volume={35},
  pages={12542--12553},
  year={2022},
  url={https://dl.acm.org/doi/abs/10.5555/3600270.3601181},
}

@inproceedings{zhou2023hacman,
  title={HACMan: Learning Hybrid Actor-Critic Maps for 6D Non-Prehensile Manipulation},
  author={Zhou, Wenxuan and Jiang, Bowen and Yang, Fan and Paxton, Chris and Held, David},
  booktitle={Conference on Robot Learning},
  pages={241--265},
  year={2023},
  organization={PMLR},
  url={https://hacman-2023.github.io/},
}

@inproceedings{foundationpose,
  title={Foundationpose: Unified 6d pose estimation and tracking of novel objects},
  author={Wen, Bowen and Yang, Wei and Kautz, Jan and Birchfield, Stan},
  booktitle={Proceedings of the IEEE/CVF Conference on Computer Vision and Pattern Recognition},
  pages={17868--17879},
  year={2024},
  url={https://arxiv.org/abs/2312.08344},
}

@article{mujocoplayground,
  title={Mujoco playground},
  author={Zakka, Kevin and Tabanpour, Baruch and Liao, Qiayuan and Haiderbhai, Mustafa and Holt, Samuel and Luo, Jing Yuan and Allshire, Arthur and Frey, Erik and Sreenath, Koushil and Kahrs, Lueder A and others},
  journal={arXiv preprint arXiv:2502.08844},
  year={2025},
  URL = {https://arxiv.org/abs/2502.08844},
}

@article{ren-object-centric,
  title={Object-centric kinodynamic planning for nonprehensile robot rearrangement manipulation},
  author={Ren, Kejia and Wang, Gaotian and Morgan, Andrew S and Kavraki, Lydia E and Hang, Kaiyu},
  journal={IEEE Transactions on Robotics},
  year={2025},
}

@ARTICLE{dkl-control,
  author={Reed, Robert and Laurenti, Luca and Lahijanian, Morteza},
  journal={IEEE Control Systems Letters}, 
  title={Promises of Deep Kernel Learning for Control Synthesis}, 
  year={2023},
  url={https://ieeexplore.ieee.org/abstract/document/10349682},
}

\end{document}